# Do Bayesian Neural Networks Improve Weapon System Predictive Maintenance?

*Michael Potter, Naval Surface Warfare Center – Corona Division
*Miru Jun, Naval Surface Warfare Center – Corona Division

* denotes equal contribution

## ABSTRACT

We implement a Bayesian inference process for Neural Networks to model the time to failure of highly reliable weapon systems with interval-censored data and time-varying covariates. We analyze and benchmark our approach, LaplaceNN, on synthetic and real datasets with standard classification metrics such as Receiver Operating Characteristic (ROC) Area Under Curve (AUC) Precision-Recall (PR) AUC, and reliability curve visualizations.

## 1 INTRODUCTION

Recent advances in deep learning software such as PyTorch have enabled rapid experimentation and prototyping of deep learning models in the research community [1]. However, deep learning models are known for overconfident predictions and poor calibration, which leads to poor generalization of out-of-distribution (OOD) data and unobserved data [2]. Another issue with deep learning is that the experiment data must be partitioned into training, validation, and test sets for early stopping and hyperparameter tuning, resulting in less data representation and availability for development.

To combat these issues, there has been a resurgence in literature incorporating Bayesian inference in deep learning [3]. Bayesian inference does not typically require a validation dataset, and the traditional log marginal likelihood objective can be optimized with respect to the continuously differentiable neural network hyperparameters. In particular, the Laplace approximation to approximate the Neural Network parameter posterior distribution centered at the Maximum A Posteriori (MAP) point estimates has shown success [4]. The covariance matrix from the Laplace approximation for deep learning has become computationally feasible by further approximations such as the Generalized Gauss-Newton (GGN) method, and PyTorch's LAPACK library [1, 5]. It has been shown empirically that when using the GGN approximation, the model used in the posterior predictive distribution should be linearized. The Laplace approximation in combination with the linearization of the deep learning model introduces model uncertainty and reduces overfitting [6].

For weapon systems in the Navy, most of the current failure time reliability modeling practices employ only a conditional Weibull distribution, based on the weapon system age at test time and the weapon system age (age) at the last time of test (agelt). This approach lacks the extra information on individual weapon system characteristics. A recent method introduced the Weibull-Cox Bayesian Neural Network tested on several weapon systems, albeit requiring a held-out validation set [7]. Moreover, while understanding the population reliability trends via a Weibull distribution is informative, this formulation does not incorporate individual weapon system characteristics to acquire personalized reliability curves. Incorporating other important features, such as weapon manufacturers, storage duration in detrimental environments, and weapon system configurations may lead to drastically different reliability estimates at any given point in time. Finally, despite the abundant recent literature around deep learning, with relational datasets such as weapon system data, non-deep learning methods typically prevail in performance [8].

Motivated by these observations, our contributions comprise benchmarking several Navy weapon systems with improved modeling capabilities:

1. We incorporate new and informative weapon system features in the intensity function of a Weibull regression model, as a neural network function or a linear function of the covariate, without the need for a validation set.

2. We propose a simple Bayesian Neural Network for Weibull Regression via the Laplace approximation, the GGN approximation, and a local linearization of the neural network in the posterior predictive distribution.

3. Considering our relational data, we further propose a high-dimensional Bayesian linear Weibull Regression via mixed/continuous Hamiltonian Monte Carlo inference methods.

4. Incorporating the advancements of probabilistic programming languages such as NumPyro, we perform Bayesian inferences of high dimensional posterior distributions for Weibull regression models using several different priors for comparison [9, 10].

## 2 PROBLEM FORMULATION AND DATASETS

We benchmark our methodology on 7 datasets shown in

Table 1, including two weapon system datasets. For the Banana, Moon, Banana 2, Moon 2, and Heart Failure we assume the data is non-repairable, and thus the interpretation of equations section 3 becomes a Weibull-Cox formulation.

*Table 1: Dataset Descriptive Statistics (on the training partition)*

| Dataset | # of Samples | # of Features | % Failed |
|---|---|---|---|
| Banana | 1579 | 2 | 0.335 |
| Moon | 1546 | 2 | 0.323 |
| Banana 2 | 1052 | 2 | 0.504 |
| Moon 2 | 1071 | 2 | 0.467 |
| Heart Failure | 1140 | 11 | 0.675 |
| Weapon System 1 | 1000s | >10 | – |
| Weapon System 2 | 1000s | >50 | – |

We partition each dataset into train and test sets by the unique identifier of an item for a set of binary test results, i.e., pass or fail labels associated with the respective item. For example, weapon system "A" may have 3 tests, weapon system "B" may have 10 and weapon system "C" may have 5 tests. All tests from weapon system "A" and weapon system "B" will belong to the train set, while all tests from weapon system "C" will belong to the test set. There are $J$ unique items, and each unique item has $|I_j|$ test results. Therefore, each dataset contains $N = \sum_j^M |I_j|$ test intervals (datapoints), and follows the relational form:

$$\mathcal{D} = \left( X \in \mathbb{R}^{N \times F}, y \in \{0,1\}^{N \times 1}, t_{agelt} \in \mathbb{R}^{N \times 1}, t_{age} \in \mathbb{R}^{N \times 1} \right)$$

$X$ denotes the samples of the dataset, where each sample has $F$ covariates. Each binary scalar element of label vector $y$ is a functionality test, where 1 denotes a failure and 0 denotes a pass. Each element of $t_{age}$ and $t_{agelt}$ represent the total age at the test interval $i$ (right side of the interval) denoted by $t_{age,ji}$ and age at the last test time denoted by $t_{agelt,ji}$ (left side of interval) for a unique item $j \in [1, J]$, respectively.

We explain the details of each dataset below.

2.1 *Synthetic Nonlinear Data I*

We consider two synthetic datasets: the moon and the banana datasets [11, 12]. Each dataset is a binary classification dataset, which we convert into an interval-censored reliability dataset. We simulate the time to failure for each datapoint within a given binary class with rate $\lambda_{class}$ and shape $k_{class}$, and then move a disjoint sliding window across time as the testing interval. Pseudocode of data generation is shown below:

$\lambda = [0.1, 0.5]; k = [2.5, 3.0]$
$for\ class, (\lambda_{class}, k_{class})\ in\ enum((\lambda, k)):$
$\quad X_{class} = X[y = class]$
$\quad u \sim U(0, 1)^N$

$$t_{fail} \sim \frac{(-log(1-u))^{\frac{1}{k_{class}}}}{\lambda_{class}}$$

$chunk\ t_{fail}\ into\ interval\ observations$
$for\ each\ interval\ i, y_i = 1\ if\ t_{agelt} < t_{fail} < t_{age}\ else\ 0$

An example is shown in Figure 1 with unique items A, B, C, and D which fail at years 3, 3.8, 7, and 10, respectively. We move a disjoint sliding window of size 2 years across time and test the item at the right-hand side of the interval. The dataset is then comprised of the tuples shown in Table 2.

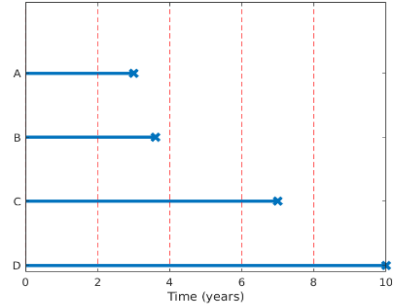

*Figure 1: Converting Time to Failure Data to Interval Censored Data*

*Table 2: Datapoints from Converted Time to Failure Dataset*

| [0,2] | [2,4] | [4,6] | [6,8] | [8,10] |
|---|---|---|---|---|
| $(X_A, 0, 0, 2)$ | $(X_A, 1, 2, 4)$ | | | |
| $(X_B, 0, 0, 2)$ | $(X_B, 1, 2, 4)$ | | | |
| $(X_C, 0, 0, 2)$ | $(X_C, 0, 2, 4)$ | $(X_C, 0, 4, 6)$ | $(X_C, 1, 6, 8)$ | |
| $(X_D, 0, 0, 2)$ | $(X_D, 0, 2, 4)$ | $(X_D, 0, 4, 6)$ | $(X_D, 0, 6, 8)$ | $(X_D, 1, 8, 10)$ |

2.2 *Synthetic Nonlinear Data II*

Similar to synthetic nonlinear data I, we use the moon and banana datasets to simulate interval censored time to failure data, with a modification in modeling the rate. In this setting, the failure rate is a continuous function of the features:

$k = [2.5, 3.0]$
$for\ class, (k_{class})\ in\ enum(k):$
$\quad X_{class} = X[y = class]$
$\quad u \sim U(0, 1)^N$
$\quad if\ class\ 0, \lambda = e^{-0.2 \times x_1 - 0.15 \times x_2 - 1}$
$\quad if\ class\ 1, \lambda = e^{-0.5 \times x_1^2 - 0.15 \times x_2^2} + 0.8$

$$t_{fail} \sim \frac{(-log(1-u))^{\frac{1}{k_{class}}}}{\lambda}$$

$chunk\ t_{fail}\ into\ interval\ observations$
$for\ each\ interval\ i, y_i = 1\ if\ t_{agelt} < t_{fail} < t_{age}\ else\ 0$

2.3 *Heartfailure Data*

299 heart failure patients were admitted to the Institute of

Cardiology and Allied hospital Falsalabad-Pakistan between April-December (2015), where heart failure was observed upon admittance time [13]. If a patient died, the time to failure with respect to the heart failure admittance was recorded. We move a disjoint sliding window across time as the testing interval to create an interval-censored dataset similar to the synthetic nonlinear data II with continuous rate. This is analogous to a heart failure patient checking in with the hospital every month.

### 2.4 Weapon System Data

We consider two weapon system datasets, which are interval-censored by the nature of the weapon system lifecycle. The weapon system is taken out of the fleet for maintenance and storage (inbound test), then subsequently injected back into the fleet (outbound test) for some period of time, as in Figure 2. This process is repeated, and we do not know when the weapon system failed between the most recent outbound and the next inbound test. We further assume that no more than one failure may occur between $t_{agelt}$ and $t_{age}$.

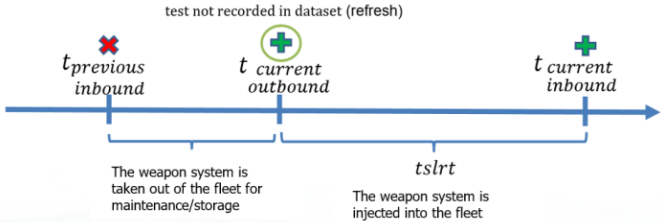

*Figure 2: Weapon System Lifecycle, where the green plus indicates a passed test, and a red x denotes a failed test*

A dataset sample $(x_i, y_i, t_{agelt,ji}, t_{age,ji})$ denotes the measured features in the interval $[t_{agelt,ji}, t_{age,ji}]$ and test result of a specific weapon system at time $t_{age,ji}$. More details of the weapon system may not be disclosed due to the sensitive nature of the Navy weapon systems.

### 3 MODEL DEFINITIONS

We assume that a weapon system is a repairable item that follows a Non-Homogenous Poisson Process (NHPP) for failures, with immediate repair time for simplicity. For an NHPP, the number of failures in any interval is also a Poisson random process:

$$N(t_{age}) - N(t_{agelt}) \sim Poisson\left(\int_{t_{agelt}}^{t_{age}} h(\tau, x(\tau))d\tau\right) \quad (1)$$

The probability of no failures in an interval then follows:

$$P[N(t_{age}) - N(t_{agelt}) = 0 | x(t)] = e^{-\int_{t_{agelt}}^{t_{age}} h(\tau, x(\tau))d\tau} \quad (2)$$

We parameterize the intensity function as:

$$h(t, x(t)) = e^{g_\theta(x(t))}(\lambda t)^{k-1} k\lambda \quad (3)$$

defined by the shape ($k$), rate ($\lambda$), and function parameters $\boldsymbol{\theta}$.

Given that an item $j$ received maintenance/inspection at $t_{agelt,ji}$, the reliability of the weapon system and the complement Cumulative Distribution Function (CDF) of the inter-arrival times ($T_{ji}$) to the next failure are then given by:

$$P[T_{ji} > t_{age,ji} | x_j(t), \lambda, k, \boldsymbol{\theta}] = e^{-\int_0^{t_{age,ji}} e^{g_\theta(x(\tau))}(\lambda\tau)^{k-1} k\lambda \, d\tau} \quad (4)$$

$$P[T_{ji} > t_{age,ji} | x_j(t), T_{ji} > t_{agelt,ji}, \lambda, k, \boldsymbol{\theta}]$$
$$= e^{-\int_{t_{agelt,ji}}^{t_{age,ji}} e^{g_\theta(x(\tau))}(\lambda\tau)^{k-1} k\lambda \, d\tau} \quad (5)$$

We note Eq. 4 and 5 follow Weibull Cox and conditional Weibull Cox models, respectively.

The time-varying covariates, $x_j(t)$, are constant within each interval $[t_{agelt,ji}, t_{age,ji}]$ when the weapon system is deployed in the fleet. Therefore, the $e^{g_\theta}$ term can be pulled out of the time dependent integral:

$$P[T_{ji} > t_{age,ji} | x_{ji}, T_{ji} > t_{agelt,ji}, \lambda, k] =$$
$$e^{-e^{g_\theta(x_{ji})} \int_{t_{agelt,ji}}^{t_{age,ji}} (\lambda\tau)^{k-1} k\lambda \, d\tau} \quad (6)$$

### 3.1 Linear Weibull Regression

We first consider the posterior distribution of the linear Weibull Regression model parameters:

$$f_{R,K,\Theta|D}(r, k, \boldsymbol{\theta}|\mathcal{D})$$
$$\propto f(\mathcal{D}|r, k, \boldsymbol{\theta}) f_R\left(R|R_{t_{fix}}\right) f_K(k) f_\Theta(\boldsymbol{\theta}) \quad (7)$$

where $R_{t_{fix}}$ is an initial reliability estimate at a given time $t_{fix}$ that contractors provide from accelerated life testing. The prior distribution and the posterior distribution for the rate parameter is related to $R$ and $K$ via change of random variables (which may not be specified due to sensitive nature of the Navy):

$$\Lambda = \zeta(K, R, t_{fix}) \quad (8)$$

From here on, we will drop the model parameters and covariates in any conditional probability descriptions for conciseness, unless particularly needed.

The likelihood function for interval censored weapon system failures comes from the NHPP inter-arrival observations:

$$f(y|t_{age}, t_{agelt}, \boldsymbol{\theta}, \lambda, k, X)$$
$$= \prod_{j=1}^{M} \prod_{\{i \in |I_j|\}} P[N_j(t_{age,ji}) - N_j(t_{agelt,ji}) = 0|x_{ji}]^{1-y_{ji}} \times$$
$$(1 - P[N_j(t_{age,ji}) - N_j(t_{agelt,ji}) = 0|x_{ji}])^{y_{ji}} \quad (9)$$

In terms of the Weibull-Cox model conditional complement CDF, the likelihood has the form:

$$f(y|t_{age}, t_{agelt}, \boldsymbol{\theta}, \lambda, k, X)$$
$$= \prod_{j=1}^{M} \prod_{\{i \in |I_j|\}} P[T_{ji} > t_{age,ji} | x_{ji}, T_{ji} > t_{agelt,ji}, \lambda, k]^{1-y_{ji}} \times$$
$$(1 - P[T_{ji} > t_{age,ji}|x_{ji}, T_{ji} > t_{agelt,ji}, \lambda, k])^{y_{ji}} \quad (10)$$

The log-likelihood is therefore:

$$\log f(y|t_{age}, t_{agelt}, \boldsymbol{\theta}, \lambda, k, X)$$
$$= \sum_{j=1}^{M} \sum_{\{i \in |I_j|\}} (1 - y_{ji}) \log P[T_{ji} > t_{age,ji} | x_{ji}, T_{ji} > t_{agelt,ji}, \lambda, k] + y_{ji} \log P[T_{ji} > t_{age,ji}|x_{ji}, T_{ji} > t_{agelt,ji}, \lambda, k]$$
$$\quad (11)$$

We experiment with two different priors for the covariate coefficients $\boldsymbol{\Theta}$: the absolutely continuous spike and slab prior and the multivariate normal distribution.

When we intend for Bayesian Variable Selection (BVS) to pinpoint possible causal features for program sponsors, we use the absolutely continuous spike and slab prior [14]. We use unimodal Normal distributions as the spike and slab components with variances $\tau_f^2$ and $v_f^2$, respectively (Normal Mixture Inverse Gamma) NMIG formulation where $\frac{\tau_f^2}{v_f^2} \gg 1$) [15]. [16] has shown model selection for linear models using covariates with posterior inclusion probability (PIP) greater than 0.5 leads to optimal posterior predictive distributions. The multivariate normal distribution prior alternative is specified by **0** mean and a diagonal covariance matrix with variance $v^2$. We set $v^2 = 1$ as the covariates are standard normalized during preprocessing.

### 3.1.3 Baseline Weibull Regression

We consider the Weibull regression without covariates as the baseline model to benchmark against, as this is the current methodology of the Navy for most weapon systems. The posterior distribution in Eq. 7 is modified to remove the covariate coefficients **Θ**:

$$f_{R,K|D}(r, k|\mathcal{D}) \propto f(\mathcal{D}|r, k) f_R(R|R_{t_{tfix}}) f_K(k) \quad (12)$$

where we use the same priors on the shape and rate parameters specified by Eq. 12 and 13.

### 3.2 Nonlinear Weibull Regression

For non-linear regression via a neural network, we employ a Laplace approximation approach [4]. We consider the posterior distribution of the nonlinear Weibull Regression model parameters: shape ($K$) and Neural Network weights (**Θ**):

$$f_{K,\Theta|D}(k, \theta|\mathcal{D}) \propto f(\mathcal{D}|k, \theta) f_K(k) f_\Theta(\theta) \quad (13)$$

In doing so, we do not incorporate the initial contractor reliability through $R$, as this is beyond the scope of the new modeling approach via Laplace approximation.

Denoting $\boldsymbol{\phi} = [\boldsymbol{\theta}, k]$, the Neural Network $g_\phi$ learns the rate and shape parameters:

$$\boldsymbol{z}_{ji} = [\lambda_{ji}, k] = g_\phi(\boldsymbol{x}_{ji}) \quad (14)$$

, where $k$ is a model parameter (and not a function of $\boldsymbol{x}_{ji}$). The mean link function follows a conditional Weibull-Cox model (c.f. Section 3):

$$p_{ji}(\lambda_{ji}, k; t_{age,ji}, t_{agelt,ji}) = e^{-\lambda_{ji}^k e^{\lambda(t_{age,ji}^k - t_{agelt,ji}^k)}} \quad (15)$$

We utilize the likelihood formulation from Eq. 8 and 11. The prior distribution on the model parameters follows a normal distribution, where $v$ is post-fit calculated from log marginal likelihood optimization following [4].

#### 3.2.1 MAP

We first find the maximum a posteriori (MAP) point estimate $\boldsymbol{\phi}^* = \underset{\phi}{\operatorname{argmax}} \ell(\boldsymbol{\phi}, D)$, where $\ell(\boldsymbol{\phi}, D)$ denotes the unnormalized log posterior distribution:

$$\ell(\boldsymbol{\phi}, \mathcal{D}) = \log f(\boldsymbol{y}|\boldsymbol{t}_{age}, \boldsymbol{t}_{agelt}, \boldsymbol{\theta}, k, \boldsymbol{X}) + \log f_\Phi(\boldsymbol{\phi}) \quad (16)$$

#### 3.2.2 Laplace Approximation

The Laplace approximation with the Generalized Gauss-Newton (GGN) covariance matrix approximation approximates the posterior distribution of the model parameters by a Gaussian distribution centered at the MAP estimate:

$$q_\Phi(\boldsymbol{\phi}) = \mathcal{N}(\boldsymbol{\phi}^*, \boldsymbol{\Sigma}_{GGN}) \quad (17)$$

The GGN approximation to the full covariance matrix given by $\Sigma_{GGN}$ is expressed as a function per data point:

$$\boldsymbol{\Sigma}_{GGN} = \left( \sum_{j \in [1,M]} \sum_{i \in I_j} \boldsymbol{J}_{\phi^*}(\boldsymbol{x}_{ji})^T \boldsymbol{H}(y_{ji}; \boldsymbol{z}_{ji}) \boldsymbol{J}_{\phi^*}(\boldsymbol{x}_{ji}) + \frac{1}{v^2} \boldsymbol{I} \right)^{-1} \quad (18)$$

where we denote $\rho = \frac{1}{v^2}$ as the precision. We express the Jacobian of the Neural Network parameters and Hessian of the log mean link function as $J_\phi(\boldsymbol{x}_{ji}) = \frac{g_\phi(\boldsymbol{x}_{ji})}{\partial \phi} \in \mathbb{R}^{2 \times P}$ and $H(y_{ji}; \boldsymbol{z}_{ji}) = -\nabla^2_{z_{ji} z_{ji}} \log P[y_{ji}|\boldsymbol{z}_{ji}] \in \mathbb{R}^{2 \times 2}$, respectively.

#### 3.2.3 Posterior Predictive

We obtain a Generalized Linear Model (GLM) posterior predictive equation by linearizing the Neural Network following [6]:

$$P_{GLM}[y^{new}|x^{new}, \mathcal{D}] = \mathbb{E}_{q_\Phi} \left[ P[y^{new}|g_{\phi^*}^{lin}(x^{new}, \boldsymbol{\phi})] \right]$$
$$\approx \frac{1}{S} \sum_{\phi_s \sim q_\Phi} P[y^{new}|g_{\phi^*}^{lin}(x^{new}, \boldsymbol{\phi}_s)] \quad (19)$$

where the linearized Neural Network function is:

$$g_{\phi^*}^{lin}(x, \boldsymbol{\phi}) = g_{\phi^*}(x) + \nabla_\phi g_\phi(x) \big|_{\phi=\phi^*}^T (\boldsymbol{\phi} - \boldsymbol{\phi}^*) \quad (20)$$

We denote the non-linearized version of the Posterior Predictive for the Neural Network as Bayesian Neural Network (BNN) [6].

## 4 MODEL INFERENCE

### 4.1 No U-Turn Sampler

We utilize the No U-Turn Samplers (NUTS) kernel for Markov Chain Monte Carlo (MCMC) sampling towards posterior inference of the Linear Weibull Regression parameters and the baseline model parameters [18]. The NUTS kernel hyper parameters are a burn-in of 20000 samples for per each of 4 chains and a target acceptance probability of 0.99. We ensure that the convergence diagnostic is close to 1, the effective number of samples is at least 1000 for each parameter, and the number of divergences and the Monte Carlo standard error are small.

### 4.2 ADAM Optimizer

When training the Neural Network to find the MAP estimate (Section 3.2.1), we use batch gradient descent with the ADAM optimizer [18]. A batch of data, denoted as $\mathcal{D}_B$, is randomly sampled without replacement from the dataset $B$ times. The objective function (Eq. 15) with the incorporation of batch size is:

$$\log f_{K,\boldsymbol{\theta}|D}(k,\boldsymbol{\theta}|\mathcal{D}) \propto \frac{N}{B}\log f(\mathcal{D}_B|k,\boldsymbol{\theta}) + \log f_K(k) + \log f_{\boldsymbol{\theta}}(\boldsymbol{\theta}) \quad (21)$$

We choose a large amount of epochs with a small batch size to ensure that gradient descent converges to the MAP estimate.

## 5 RESULTS

### 5.1 Hyperparameter Search

For each dataset, we discover the optimal combination of hyperparameters for the LaplaceNN by performing a grid search. The hyperparameters searched are combinations of precision, batch size, number of neurons in 1 hidden layer, learning rate, epochs and whether or not coordinate descent was used. We use the log marginal likelihood values of the training dataset to select the optimal hyperparameters.

### 5.2 Metric Benchmarking

For the non-weapon system datasets, we visually check that the parametric models show similar population trends to the Kaplan Meier curves. We are unable to compute the Kaplan Meier survival functions for Weapon System 1 and Weapon System 2 due to complications with the data, such as missing measurement recordings and non-reliable unique weapon system identifiers. The overlapped plots can be found in Figures 3 – 6.

For each dataset, we compare each model against the baseline model (c.f. section 3.1.3) with standard predictive metrics such as Receiver Operating Characteristic Curve (ROC) Area Under Curve and Precision-Recall (PR) AUC (Tables 1-2). For Weapon System 1 and Weapon System 2 datasets, the percentage increase from the Baseline model metrics is given instead of the metric values due to the sensitive nature of the data. The best result of every dataset is bordered, while the second best is underlined, where higher values indicate better performance. We find that each proposal method is usually at least as good as the baseline, with some results showing marked improvements.

*Table 3: ROC-AUC Results, border is best result, underline second*

| Model / Dataset | Baseline | Full MCMC | Discrete Spike Slab MCMC | BNN | Laplace NN |
|---|---|---|---|---|---|
| Banana | 0.374 | 0.384 | 0.379 | 0.483 | 0.597 |
| Moon | 0.359 | 0.568 | 0.566 | 0.602 | 0.674 |
| Banana 2 | 0.556 | 0.578 | 0.595 | 0.678 | 0.778 |
| Moon 2 | 0.516 | 0.703 | 0.704 | 0.676 | 0.804 |
| Heartfailure | 0.138 | 0.134 | 0.195 | 0.130 | 0.130 |
| Weapon System 1 | – | +22.22% | +28.6% | +22.08% | +22.08% |
| Weapon System 2 | – | +16.6% | +4.58% | +18.12% | +18.12% |

*Table 4: PR-AUC Results, border is best result, underline second*

| Model / Dataset | Baseline | Full MCMC | Discrete Spike Slab MCMC | BNN | Laplace NN |
|---|---|---|---|---|---|
| Banana | 0.544 | 0.546 | 0.544 | 0.662 | 0.742 |
| Moon | 0.54 | 0.728 | 0.728 | 0.76 | 0.812 |
| Banana 2 | 0.550 | 0.582 | 0.583 | 0.691 | 0.765 |
| Moon 2 | 0.556 | 0.741 | 0.740 | 0.716 | 0.829 |
| Heartfailure | 0.641 | 0.670 | 0.697 | 0.712 | 0.712 |
| Weapon System 1 | – | +8.71% | +8.22% | +8.07% | +8.07% |
| Weapon System 2 | – | +13.40% | +11.55% | +10.72% | +10.72% |

### 5.3 Reliability Curves

The reliability curves from every model for each dataset were visualized with the 80% reliability credible intervals to create plots (Figures 3 – 6).

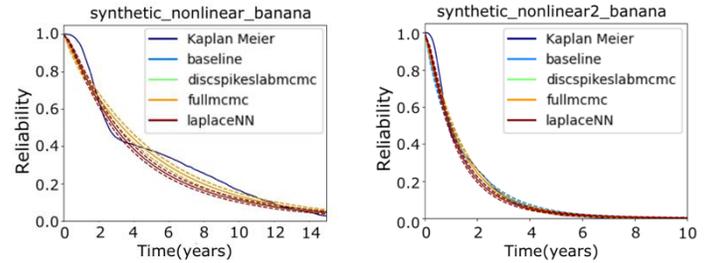

*Figure 3: Reliability Curves and 80% Reliability Credible Intervals for Each Model Trained on Banana Dataset*

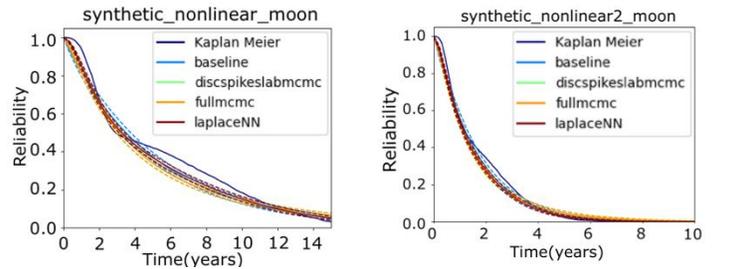

*Figure 4: Reliability Curves and 80% Reliability Credible Intervals for Each Model Trained on Moon Dataset*

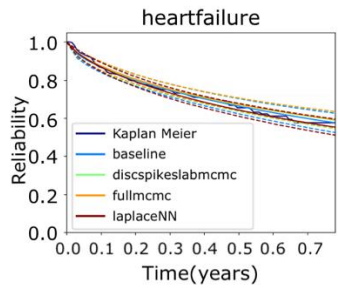

*Figure 5: Reliability Curves and 80% Reliability Credible Intervals for Each Model Trained on Heartfailure Dataset*

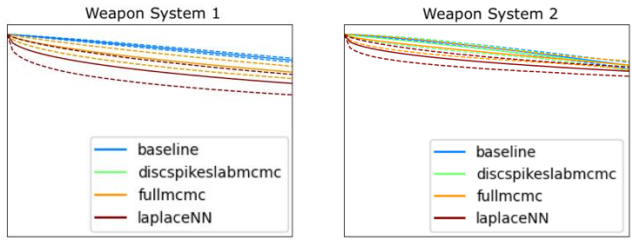

*Figure 6: Reliability Curves and 80% Reliability Credible Intervals for Each Model Trained on Weapon Systems Datasets*

## CONCLUSION

We benchmark the proposed Bayesian Neural Network and standard linear Bayesian models for interval-censored data with time-varying (and non-time-varying) covariates. The LaplaceNN model drastically improves the ROC-AUC and the PR-AUC for the synthetic datasets where the intensity function is a nonlinear function of data. However, for the heartfailure and weapon system datasets, using the log marginal likelihood selected a linear model (0 hidden layers). Therefore 0 hidden layers is a Laplace approximation with a full covariance matrix (linear model). We believe this is because the true intensity function is linear with respect to the features collected, and therefore we did not overfit with a Neural Network. Furthermore, [13] used a linear Cox-proportional hazard model. As expected, unless the features measured for the weapon systems are highly nonlinear or unstructured (such as images and text) the benefit of neural networks is minimal to none. Even though we have shown that current data available for weapon systems show minimal improvement with neural networks, in the future, larger and feature-rich datasets planned for collection will likely benefit from our AI/ML approach.

*BIOGRAPHIES*

Michael L. Potter, MS
Department Acquisition Readiness 43
Naval Surface Warfare Center - Corona
1999 Fourth St
Norco, California 92860 USA

e-mail: potter.mi@coe.northeastern.edu

Michael Potter has worked as an Electronics Engineer at the Naval Surface Warfare Center – Corona Division for over two years developing Machine Learning models. He has earned his bachelors and masters degree in Electrical and Computer Engineering at Northeastern University in 2020, and his second masters degree at the University of California Los Angeles in Electrical and Computer Engineering in 2022. He is an incoming PhD candidate at Northeastern University under Dr. Deniz Erdoğmuş of the Cognitive Systems Laboratory (CSL).

Miru D. Jun, BS Candidate
Department Acquisition Readiness 43
Naval Surface Warfare Center - Corona
1999 Fourth St
Norco, California 92860 USA

e-mail: miru.d.jun.ctr@us.navy.mil

Miru Jun is currently an intern at the Naval Surface Warfare Center – Corona Division working on various reliability projects. She is studying at the University of Southern California on track to earn a bachelors and a masters degree in Computer Science with a graduation date of 2025.

# IEEE Copyright Notice